\documentclass[11pt,a4paper]{article}
\usepackage{emnlp2020}
\usepackage{times}
\usepackage{algorithmic}
\usepackage[ruled,vlined, linesnumbered]{algorithm2e}
\usepackage{amssymb,amsmath,amsthm,enumitem}

\usepackage{hyperref}
\hypersetup{colorlinks,allcolors=black}

\usepackage{svg}
\usepackage{graphicx}
\usepackage{bm}
\usepackage{bbm}
\usepackage{enumitem}
\usepackage{textcomp}
\usepackage{multirow}
\usepackage{flushend}
\usepackage{tikz}

\usepackage[toc,page,header]{appendix}

\newcommand{\cF}{\mathcal{F}}

\newcommand{\modelname}{\qdst\xspace}

\newcommand{\mysection}[1]{{\noindent{\textbf{#1}.}\xspace}}

 \usepackage{amsthm}
\theoremstyle{definition}

\usepackage{lipsum}

\usepackage{titlesec}

\titlespacing*{\section}
{0pt}{1pt}{1pt}
\titlespacing*{\subsection}
{0pt}{1pt}{1pt}
\usepackage{pifont}
\newcommand{\cmark}{\ding{51}}%
\newcommand{\xmark}{\ding{55}}%

\newcommand{\qdst}{QDS-Transformer}

\definecolor{midnightgreen}{rgb}{0.0, 0.29, 0.33}

\usepackage{microtype}

\aclfinalcopy 
\def\aclpaperid{3388}

\title{Long Document Ranking with Query-Directed Sparse Transformer}

\author{Jyun-Yu Jiang$^\dagger$, Chenyan Xiong$^\ddagger$, Chia-Jung Lee$^{\ddagger}$ and Wei Wang$^\dagger$\\
  $^\dagger$Department of Computer Science, University of California, Los Angeles, USA \\
  $^\ddagger$Microsoft Research AI, Redmond, USA \\
  {\small \texttt{\{jyunyu,weiwang\}@cs.ucla.edu, \{chenyan.xiong,cjlee\}@microsoft.com}} \\}

\date{}
\begin{document}
\maketitle
\begin{abstract}
The computing cost of transformer self-attention often necessitates breaking long documents to fit in pretrained models in document ranking tasks.
In this paper, we design Query-Directed Sparse attention that induces IR-axiomatic structures in transformer self-attention. 
Our model, QDS-Transformer, enforces the principle properties desired in ranking:
local contextualization, hierarchical representation, and query-oriented proximity matching, while it also enjoys efficiency from sparsity.
Experiments on one fully supervised and three few-shot TREC document ranking benchmarks demonstrate the consistent and robust advantage of QDS-Transformer over previous approaches, as they either retrofit long documents into BERT or use sparse attention without emphasizing IR principles. 
We further quantify the computing complexity and demonstrates that our sparse attention with TVM implementation is twice more efficient than the fully-connected self-attention.  
All source codes, trained model, and predictions of this work are available at \url{https://github.com/hallogameboy/QDS-Transformer}.
\end{abstract}

\section{Introduction}
\label{section:introduction}

Pre-trained Transformers such as BERT~\cite{devlin2019bert} effectively transfer language understanding to better relevance estimation in many search ranking tasks~\cite{nogueira2019passage, nogueira2019multi, yang2019simple}.
Nevertheless, the effectiveness comes at the quadratic cost $O(n^2)$ in computing complexity corresponds to the text length $n$, prohibiting its direct application to long documents.
Prior work adopts quick workarounds such as document truncation or splitting-and-pooling to retrofit the document ranking task to pretrained transformers. 
Whilst there have been successes with careful architecture design, those 
bandit-solutions inevitably introduce information loss and create complicated system pipelines.

\begin{figure}[!t]
  \centering
  \includegraphics[width=\linewidth]{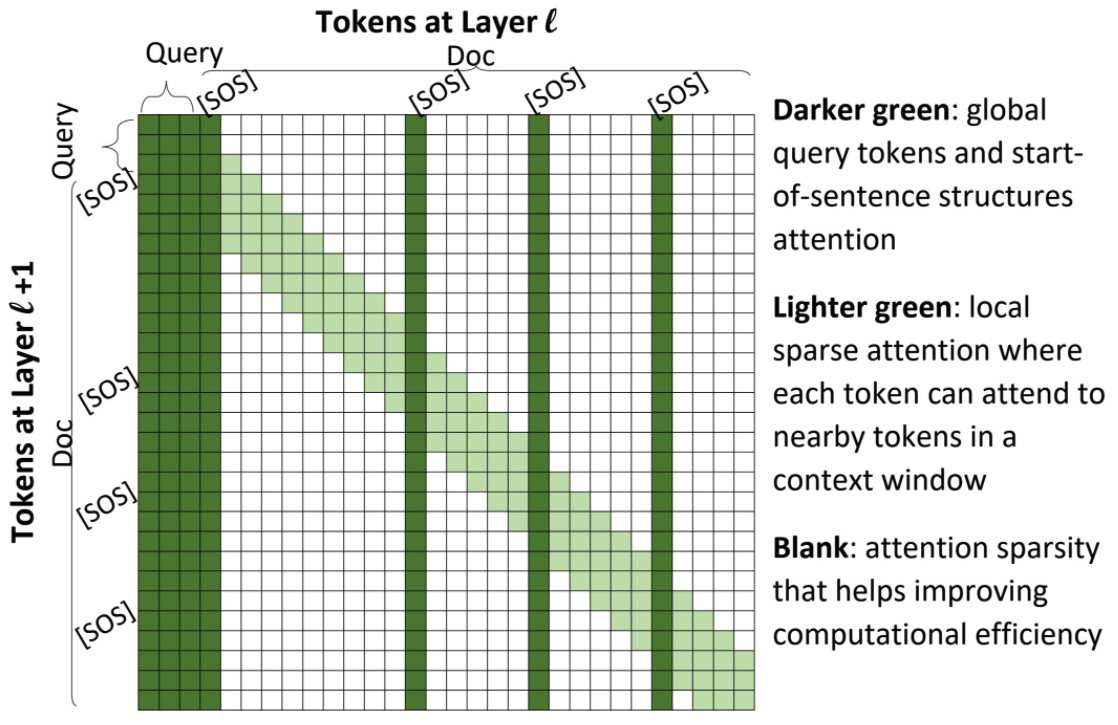}
  \caption{An example illustration of the attention mechanism used in Query-Directed Sparse Transformer.}
  \label{fig:attexp}
\end{figure}

Intuitively, effective document ranking does not require fully connected self-attention between all query and document terms.
The relevance matching between queries and documents often takes place at text segments as opposed to individual tokens~\cite{callan1994passage, jiang2019semantic}, suggesting that a document term may not need information thousands of words away~\cite{metzler2005markov, child2019sparsetransformer}, and that not all document terms are useful to calculate the relevance to the query~\cite{xiong2017end}.
The fully connected attention matrix includes many unlikely connections that create efficiency debt in computing, inference time, parameter size, and training convergence.

This paper presents Query-Directed Sparse Transformer~(\qdst{}) for long document ranking.   
In contrast to retrofitted solutions, \qdst{} fundamentally considers the desirable properties for assessing relevance by focusing on attention paths that matter. Using sparse local attention~\cite{child2019sparsetransformer}, our model removes unnecessary connections between distant document tokens.
Using global attention upon sentence boundaries, our model further incorporates the hierarchical structures within documents.
Last but not the least, we use global attention on all query terms that direct the focus to the relevance matches between query-document term pairs. 
These three attention patterns in our Query-Directed Sparse attention, as illustrated in Figure~\ref{fig:attexp}, permit global dissemination of IR-axiomatic information while keeping computation compact and essential.

In our experiments with TREC Deep Learning Track~\cite{craswell2020overview} and three more few-shot document ranking benchmarks~\cite{zhang2020selective},
\qdst{} consistently improves the standard retrofitting BERT ranking baselines (e.g., max-pooling on paragraphs) by 5\% NDCG.
It also shows gains over more recent transformer architectures that induces various sparse structures, including Sparse Transformer, Longformer, and Transformer-XH, as they were not designed to incorporate the essential information required in document ranking. 
In the meantime, we also thoroughly quantify the efficiency improvement from our query-directed sparsity, showing that with TVM support~\cite{chen2018tvm}, different sparse attention patterns lead to variant training and inference speed up, and in general \modelname{} enjoys 200\%+ speed up compared to vanilla BERT on long documents.

Our visualization also shows interesting learned attention patterns in \modelname{}. Similar to the observation on BERT in NLP pipeline~\cite{tenney2019bert}, in lower \modelname{} levels, the attention focuses more on learning the local interactions and document hierarchies, while in higher layers the model focuses more on relevance matching with the query terms.
We also show examples that QDS attention may center on the sole sentence that directly answers
the query, or may span across several sentences that cover different aspects of the query, depending on the scope of the intent; this brings the advantage of better interpretability based on sparse attention.

\section{Related Work}
\label{section:relatedwork}

Neural models have demonstrated significant advances across various ranking tasks \cite{guo2019deep_survey}. Early approaches investigated diverse ways to capture relevance between queries and documents~\cite{Guo_2016_DRMM,xiong2017end,conv_knrm, hui2017pacrr}. And recently the state-of-the-art in many text ranking tasks has been taken by BERT or other pretrained language models~\cite{devlin2019bert, nogueira2019multi, nogueira2019passage, dai2019deeper, yang2019simple,craswell2020overview}, when sufficient relevance labels are available for fine-tuning (e.g., on MS MARCO~\cite{bajaj2016ms}).

The improved effectiveness comes with the cost of computing efficiency with deep pretrained transformers, especially on long documents.
This stimulates studies investigating ways to retrofit long documents to BERT's maximum sequence length limits (512). 
A vanilla strategy is to truncate or split the documents:  
\citet{dai2019deeper} applied BERT ranking on each passage segmented from the document independently and explored different ways to combine the passage ranking scores, using the score of the first
passage (BERT-FirstP), the best passage (BERT-MaxP) (also studied in \citet{yanidst2020}), or the sum of all passage scores (BERT-SumP). 

More sophisticated approaches have also been developed to introduce structures to transformer attentions. 
Transformer-XL employs recurrence on a sequence of text pieces~\cite{dai2019transformerxl},
Transformer-XH~\cite{zhao2020transformerxh} models a group of text sequences by linking them with eXtra Hop attention paths, and Transformer Kernel Long (TKL)~\cite{hofsttter2020local} uses a 
sliding window over the document terms and matches them with the query terms using matching kernels~\cite{xiong2017end}.

On the efficiency front, \citet{kitaev2020reformer} proposed Reformer that employed locality-sensitive hashing and reversible residual layers to improve the efficiency of Transformers.
\citet{child2019sparsetransformer} introduced sparse transformers to reduce the quadratic complexity to $O(L \sqrt{L})$ by applying sparse factorizations to the attention matrix, making the use of self-attention possible for extremely long sequences. 
Subsequent work \cite{sukhbaatar2019adaptive,correia2019adaptively} leverage a similar idea in a more adaptive way.
Combining local windowed attention with a task motivated global attention, \citet{Beltagy2020Longformer} presented Longformer with an attention mechanism that scales linearly with sequence length.

\begin{figure*}[!t]
    \centering
    \includegraphics[width=\linewidth]{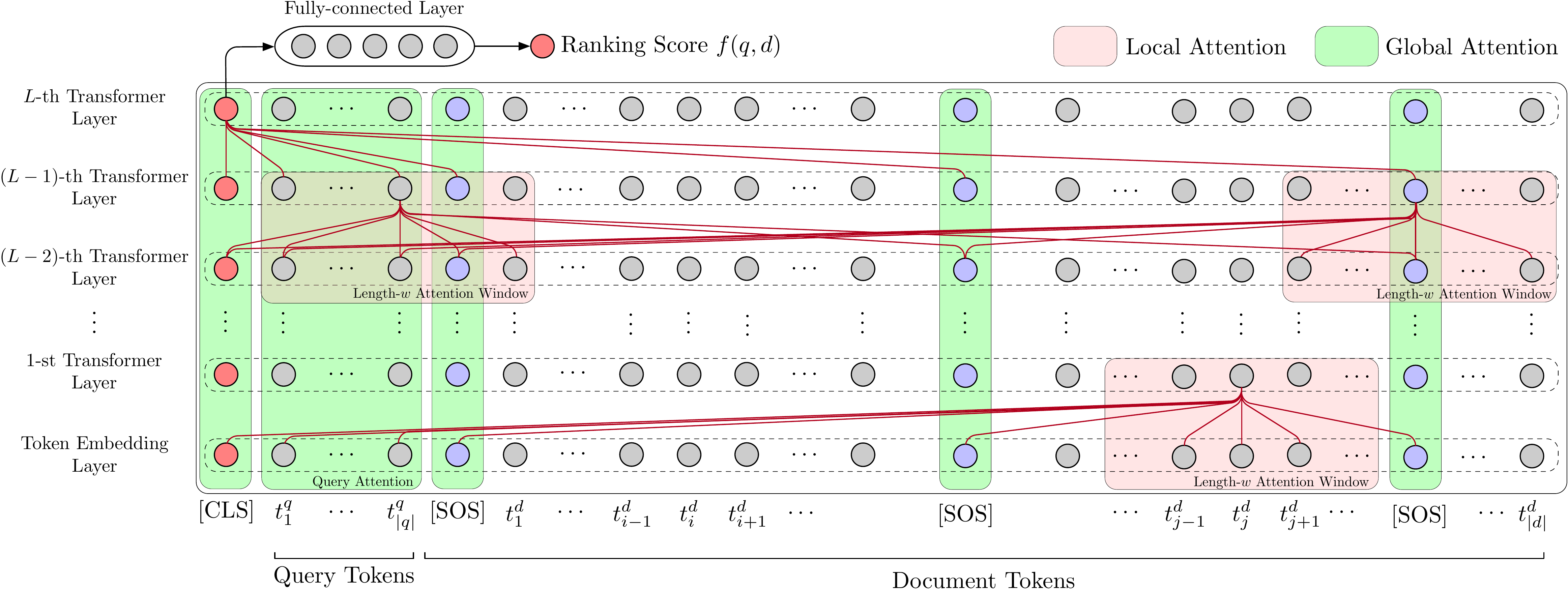}
    \caption{The overall schema of our proposed \modelname.}
    \label{fig:overview}
\end{figure*}

\section{Preliminaries on Document Ranking}
\label{section:prelim}

 Given a query $q$ and a set of candidate documents $D =\lbrace d \rbrace$, 
 the document ranking task is to produce the ranking score $f(q, d)$ for each candidate document based on their relevance to the query. 

\mysection{BERT Ranker} The standard way to leverage pretrained BERT in document ranking is to concatenate the query and the document into one text sequence,  feed it into BERT layers, and then use a linear layer on top of the last layer's [CLS] token~\cite{nogueira2019passage}:
\begin{align*}
    f(q, d) &= \text{Linear} (\text{BERT}(\text{[CLS]} \circ q \circ \text{[SEP]} \circ d)).
\end{align*}
This BERT ranker can be fine-tuned using relevance labels on $(q,d)$ pairs, as simple as a classification task, and has achieved strong performances in various text ranking benchmarks~\cite{bajaj2016ms, craswell2020overview}.

\mysection{Transformer Layer}
More specifically, let $\{t_0,t_1,...,t_i,...,t_n\}$ be the tokens in the concatenated $q$-$d$ sequence, with query tokens $t_{1:|q|} \in q$ and document tokens $t_{|q|+1:n} \in |d|$, considering special tokens being part of $q$ or $d$. The $l$-th transformer layer in BERT takes the hidden representations of previous layer ($H^{l-1}$), which is embedding for $l=1$, and produces a new $H^l$ as follows~\cite{vaswani2017attention}.
\begin{align}
    H^l &= W^F(\hat{H}^l), \label{eq.F}\\
    \hat{H}^l &= A\cdot M \cdot V^T, \label{eq.H}\\
        A & = \mathbf{1}, \label{eq.A}\\
    M & = \text{softmax}(\frac{Q \cdot K^T}{\sqrt{d_k}}), \label{eq.attM}\\
    (Q^T; K^T; V^T) &= (W^q; W^k; W^v)\cdot H^{l-1}. \label{eq.attp}
\end{align}
It first passes the previous representations through the self-attention mechanism, using three projections (Eqn.~\ref{eq.attp}), and then calculates the attention matrix between all token pairs using their query-key similarity (Eqn.~\ref{eq.attM}, as in single-head formation). The attention matrix $M$ then is used to fuse all other tokens' representation $V$, to obtain the updated representation for each position (Eqn.~\ref{eq.H}). In the end, another feed-foreword layer is used to obtain the final representation of this layer $H^l$ (Eqn.~\ref{eq.F}). 

The matrix $A$ is the $n^2$ ``adjencency'' matrix in which each entry is one if there is an attention path between corresponding positions: $A_{ij}=1$ means $t_i$ queries the value of $t_j$ using the key of $t_j$. In standard transformer and BERT, the attention paths are fully connected thus $A=\mathbf{1}$.

\mysection{Computation Complexity}
In each of the BERT layers, all the feed-forward operations (Eqn.~\ref{eq.F} and ~\ref{eq.attp}) are applied to each individual token, leading to linear complexity w.r.t. text length $n$ and the square of the hidden dimension size $dim$. The self-attention operation in Eqn.~\ref{eq.H} and~\ref{eq.attM} calculates the attention strengths upon all token pairs, leading to squared complexity w.r.t text length but linear of the hidden dimension size. 

The complexity of one transformer layer in BERT thus includes two components:
\begin{align}
   \underbrace{\mathcal{O}(\text{dim}^2n)}_\text{Feedforward} +  \underbrace{\mathcal{O}(n^2\text{dim})}_\text{Self-Attention}.
\end{align}
The hidden dimension size (dim) is 768 in BERT Base and 1024 in BERT Large~\cite{devlin2019bert}. When the text sequence is longer than 1000 or 2000 tokens, which is often the case in document ranking~\cite{craswell2020overview}, the self-attention part becomes the main bottleneck in both computation and GPU memory. 
This leads to various retrofitted solutions that adapted the document ranking tasks to standard BERT which takes at most 512 tokens per sequence~\cite{dai2019deeper, yang2019simple, yanidst2020, nogueira2019multi}.

\section{\modelname}
\label{section:method}

Recent research has shown that with sufficient training and fully-connected self-attention,  BERT learns attention patterns that capture meaningful structures in language~\cite{clark2019bertattention} or for specific tasks~\cite{zhao2020transformerxh}. However, this is not yet the case in long document ranking as computing becomes the bottleneck.

This section first presents how we overcome this bottleneck by injecting IR-specific inductive bias as sparse attention patterns. Then we discuss the efficient implementation of sparse attention.

\subsection{Query-Directed Sparse Attention}

Mathematically, inducing sparsity in self-attention is to modify the attention adjacency matrix $A$ by only keeping connections that are meaningful for the task. For document retrieval, we include two groups of informative connections as sparse adjacency matrices: \textit{local attention} and \textit{query-directed global} attention.

\subsubsection{Local Attention}
Intuitively, it is unlikely that a token needs to see another token thousands of positions away to learn its contextual representation, especially in the lower transformer layers which are more about syntactic and less about long-range dependencies~\cite{tenney2019bert}. We follow this intuition used in the Sparse Transformer~\cite{child2019sparsetransformer} and define the following local attention paths:
\begin{align}
    A_\text{local}[i, j]  =  1,  \text{iff } |i-j| \leq w/2.
\end{align}
It only allows a token to see another token in each transformer layer if the two are $w/2$ position away, with $w$ the window size. The local attention serves as the backbone for many sparse transformer variations as it provides the basic local contextual information~\cite{correia2019adaptively, sukhbaatar2019adaptive, Beltagy2020Longformer}.

\subsubsection{Query-Directed Global Attention}

The local attention itself does not fully capture the relevance matches between the query and documents. We introduce several query-directed attention patterns to incorporate inductive biases widely used in document representation and ranking.

\mysection{Hierarchical Document Structures} A common intuition in document representation is to leverage the hierarchical structures within documents, for example, words, sentences, paragraphs, and sections, and compose them into hierarchical attention networks~\cite{yang2016hierarchical}. We use a two-level word-sentence-document hierarchy and inject this hierarchical structure by adding fully connected attention paths to all the sentences.

Specifically, we first prepend a special token  [SOS] (start-of-sentence) to each sentence in the document, and form the following attention connections:
\begin{align}
    A_\text{sent}[i, j] = 1, \text{ iff } t_j = \text{[SOS]}.
\end{align}

\mysection{Matching with the Query}
For retrieval tasks, arguably the most important principle is to capture the semantic matching between queries and documents. Inducing this information is as simple as adding dedicated attention paths on query terms:
\begin{align}
    A_\text{query}[i, j] = 1, \text{ iff } t_i \in q.
\end{align}
It allows each token to see all query terms so as to learn query-dependent representations.

\subsection{Summary}
The three attention patterns together form the query-directed attention in \modelname{}:
\begin{align}
    A_\text{QDS} &= A_\text{local} \cup A_\text{sent} \cup A_\text{query} \cup A_\text{[CLS]}.
\end{align}
We also add the global attention between all other tokens and [CLS].
Keeping everything else standard in BERT and using this query-directed sparse attention ($A_\text{QDS}$) in place of the fully-connected self-attention ($A$), we obtain our \modelname{} architecture as illustrated in Figure~\ref{fig:overview}.


Interestingly, \modelname{} also resembles various effective IR-Axioms developed in past decades. For example, in QDS attention, a query term mainly focuses on the [SOS] token through $A_\text{Sent}$, while the [SOS] token recaps the proximity~\cite{callan1994passage} matches locally around it through $A_\text{Local}$.
The local attention in the query part also resembles the effective phrase matches~\cite{metzler2005markov} as the query term representations are contextualized using other query terms through $A_\text{Local}$.

\subsection{Efficient Sparsity Implementation}
Our query-directed sparse attention reduces the self-attention complexity from $\mathcal{O}(n^2\text{dim})$ to $\mathcal{O}(n \cdot \text{dim} \cdot (w+|q|+|s|))$, where the local window size $w$ and query length $|q|$ are constant to document length, and the number of sentences is orders of magnitude smaller.

However, to implement this sparsity efficiently on GPU is not that straightforward.
Naively using for-loops or masking the adjacency matrix $A$ may result in even worse efficiency than the full self-attention in common deep learning frameworks. 
An efficient implementation of sparse operations often requires customized CUDA kernels, which are inconvenient and require expertise in low-level GPU operations~\cite{child2019sparsetransformer}.
Inspired by Longformer~\cite{Beltagy2020Longformer}, we address this issue by implementing \modelname with Tensor Virtual Machine (TVM)~\cite{chen2018tvm}.
Precisely, we implement custom CUDA kernels using TVM to dynamically compile our attention map $A_\text{QDS}$ into efficiency-optimized CUDA codes.


\section{Experimental Methodologies}
\label{section:experiments}
This section discusses our experimental settings.

\mysection{TREC 2019 Deep Learning Track Benchmark}
We evaluate \modelname{} based on the document ranking task from this recent TREC benchmark~\cite{craswell2020overview}, specifically using the reranking subtask to rerank top-100 BM25 retrieved documents.
The official evaluation metric is NDCG@10 on the test set. We also report MAP on test and MRR@10 on the development set.

\mysection{Few-shot Document Ranking Benchmarks}
We then evaluate the generalization ability of \modelname{} in the few-shot setting~\cite{zhang2020selective} using TREC datasets Robust04~(RB04),  ClueWeb09-B~(CW09), and ClueWeb12-B13~(CW12), in which labels are much fewer than DL track.
Our experimental settings are consistent with prior work~\cite{zhang2020selective} in using the``MS MARCO Human Labels''. Specifically, neural rankers trained with MARCO labels are used as feature extractors to enrich TREC documents, which are then tested with five-fold cross-validation~\cite{conv_knrm}.

Table~\ref{tab:datasets} summarizes the statistics of four datasets. We describe more details about datasets and experimental settings in Appendix~\ref{appendix:datasets}.

\begin{table}[!t]
    \centering
    \resizebox{\linewidth}{!}{
    \begin{tabular}{l|r|rrr} \hline
                    & Ad-hoc  & \multicolumn{3}{c}{Few-shot (avg. over 5 folds)}\\ \cline{2-5}
                    & TREC19 DL & RB04 & CW09-B & CW12-B13 \\  \hline
    Train queries &  367,013 & 150 & 120 & 60 \\
    Train qrels &  384,597 & 186,846 & 28,278 & 17,343 \\
    \hline
    Dev queries &  5,193 & 50 & 40 & 20\\
    Dev qrels &  519,300 & 62,282  & 9,426 & 5,781\\ \hline
    Test queries &  43 & 50 & 40 & 20\\
    Test qrels & 16,258 & 62,282  & 9,426 & 5,781\\
    \hline
    \end{tabular}}
    \caption{The statistics of the experimental datasets.}
    \label{tab:datasets}
\end{table}

\mysection{Baselines} Our baselines include multiple neural IR models and the best official TREC runs of single models.
The main baselines cover:

    \setlist{nolistsep}
    \begin{itemize}[leftmargin=2.5mm,nosep]
        \item Relying on BERT models, RoBERTa (FirstP) only considers the first paragraph, while RoBERTa (MaxP) encodes short paragraphs with BERT and combines them with a max-pooling layer~\cite{dai2019deeper}.
        \item Transformer-XH~\cite{zhao2020transformerxh} retrofits data pipelines to create independent sentences which are fed into BERT models, and aggregates them with an extra-hop attention layer.
        \item TK~\cite{hofstatter2020interpretable} and TKL~\cite{hofsttter2020local} apply BERT-based kernels to estimate the relevance over document tokens with full attention.
        \item Sparse-Transformer~\cite{child2019sparsetransformer} applies length-$w$ sparse local attention windows without considering query tokens.
        \item Longformer also uses sparse local attention and adds global attention by prepending one special token respectively to the query and document, same as in their~\cite{Beltagy2020Longformer} QA setup. 
    \end{itemize}

For ad-hoc retrieval, we also consider CO-PACRR~\cite{hui2018co} which employs CNNs without using pretrained NLM (non-PLM). 
Note that IDST~\cite{yanidst2020} is not comparable because it exploits external generators for document expansion.
For the few-shot learning task, we additionally compare with SDM, RankSVM, Coor-Ascent, and Conv-KNRM as reported in previous studies~\cite{xiong2017end,conv_knrm}.
More details of the baselines can be found in Appendix~\ref{appendix:baselines}.

\mysection{Implementation Details}
\label{section:implementationdetails}
We implement all methods with PyTorch~\cite{paszke2019pytorch} and the Hugging Face transformer library~\cite{wolf2019transformers}, excluding the baselines that have previously reported their scores.
For sparse attention, we implement it using TVM with a custom CUDA kernel in PyTorch~\cite{chen2018tvm}.
Models are optimized by the Adam optimizer~\cite{kingma2014adam} with a learning rate $10^{-5}$, $(\beta_1, \beta_2) = (0.9, 0.999)$, and a dropout rate 0.1.
The dev set is used for hyperparameter tuning to decide the best model, which is then applied to the test set.
We set the maximum length of input sequences as 2,048.
The dimension of the dense layer $\cF_{\text{dense}}(\cdot)$ in relevance estimation is 768, while the local attention window size $w$  is 128.
All experiments are conducted on an Nvidia DGX-1 server with 512 GB memory and 8 Tesla V100 GPUs.
Each method is limited to access only one GPU for fair comparisons.

\begin{table}[!t]
    \centering
    \resizebox{\linewidth}{!}{
    \begin{tabular}{l|cc|c} \hline
    \multicolumn{4}{r}{\textbf{TREC Deep Learning Track Document Ranking}} \\ \hline
    \multirow{2}{*}{Method} & \multicolumn{2}{c|}{Test Set} & Dev Set  \\
    \cline{2-4}
          & NDCG@10 & MAP &  MRR@10  \\ \hline 
          
BM25 & 0.488 & 0.234 & 0.252 \\
\hline \hline \multicolumn{4}{l}{\textbf{TREC Best Models}} \\ \hline
BM25 (bm25tuned\_prf) & 0.528 & 0.386 & 0.318 \\
Trad (srchvrs\_run1) & 0.561 & 0.349 & 0.306 \\
Non-PLM (TUW19-d3-re) & 0.644 & 0.271 & 0.401 \\
BERT (bm25exp\_marcomb) & 0.646 & 0.424 & 0.352 \\
\hline \hline \multicolumn{4}{l}{\textbf{Baseline Models}} \\ \hline
CO-PACRR & 0.550 & 0.231 & 0.284 \\
TK & 0.594 & 0.252 & 0.312 \\
TKL & 0.644 & 0.277 & 0.329 \\
RoBERTa (FirstP) & 0.588 & 0.233 & 0.278 \\
RoBERTa (MaxP)  & 0.630 & 0.246 & 0.320 \\
\hline \hline \multicolumn{4}{l}{\textbf{Sparse Attention based Models}} \\ \hline
Sparse-Transformer & 0.634 & 0.257 & 0.328 \\
Longformer-QA & 0.627 & 0.255 & 0.326 \\
Transformer-XH & 0.646 & 0.256 & 0.347 \\
\hline \hline
\modelname & \textbf{0.667} & \textbf{0.278} & \textbf{0.360} \\

\hline
          
    \end{tabular}}
    \caption{The ad-hoc retrieval performance of our approach and baseline methods on the TREC-19 DL track benchmark. Note that those baselines with higher MAP scores are all full retrieval and benefited from additional data engineering like query expansion.}
    \label{tab:mainexp}
\end{table}

\section{Evaluation Results} 
This section evaluates \modelname{} in its effectiveness, attention patterns, and efficiency. We also analyze the learned query-directed attention weights and show case studies.

\subsection{Retrieval Effectiveness}
Table~\ref{tab:mainexp} summarizes the retrieval effectiveness on the TREC-19 DL benchmark. Table~\ref{tab:fewshot} shows the few-shot performance on the three TREC datasets.

\modelname consistently outperforms baseline methods on all datasets in both experimental settings.
Note that the higher MAP scores from some methods in TREC-19 DL is because they have better first stage retrieval and are not using the same reranking setting.
\modelname outperforms the best BERT-based TREC run by 3.25\% in NDCG@10 and is more effective than the concurrent sliding window approach, TKL.
Moreover, \modelname outperforms RoBERTa~(MaxP), which is the standard retrofitted method for BERT, by 6\% in NDGG@10 while also being a unified framework.

Compared with Sparse Transformers and Longformer-QA, \qdst{} provides more than 5\% improvement in nearly all datasets.
The best baseline is Transformer-XH, which creates structural sparsity by breaking a document into segments and  
introduces effective eXtra-hop attentions to jointly model the relevance of those segments. 
While these methods show competitive effectiveness especially with our TVM implementation,
\qdst{} is consistently more accurate through the query-directed sparse attention patterns in all evaluation settings.

\begin{table}[!t]
    \centering
        \resizebox{\linewidth}{!}{
    \begin{tabular}{lcccccc} \hline
         \multirow{2}{*}{\textbf{Method}}&  \multicolumn{2}{c}{\textbf{RB04}} & \multicolumn{2}{c}{\textbf{CW09}} & \multicolumn{2}{c}{\textbf{CW12}} \\ \cline{2-7}
         & NDCG & ERR & NDCG & ERR & NDCG & ERR \\ \hline
\hline \multicolumn{6}{l}{\textbf{Classical IR; Cross Validated}}          \\
SDM & 0.427 & 0.117 & 0.277 & 0.138 & 0.108 & 0.091 \\
RankSVM & 0.420 & n.a. & 0.289 & n.a. & 0.121 & 0.092 \\
Coor-Ascent & 0.427 & n.a. & 0.295 & n.a. & 0.121 & 0.095 \\
\hline \multicolumn{6}{l}{\textbf{Neural IR; Trained on MS MARCO and then Cross Validated.}} \\
Conv-KNRM & 0.427 & 0.117 & 0.287 & 0.160 & 0.112 & 0.092 \\
RoBERTa (FirstP) & 0.437 & 0.110 & 0.262 & 0.161 & 0.111 & 0.086 \\
RoBERTa (MaxP) & 0.439 & 0.114 & 0.264 & 0.162 & 0.092 & 0.074 \\
Sparse-Transformer & 0.449 & 0.119 & 0.274 & 0.173 & 0.119 & 0.094 \\
Longformer-QA & 0.448 & 0.113 & 0.276 & 0.179 & 0.111 & 0.085 \\
Transformer-XH & 0.450 & 0.123 & 0.283 & 0.179 & 0.107 & 0.080 \\
\hline
\modelname & \bf 0.457 & \bf 0.126 & \bf 0.308 & \bf 0.191 & \bf 0.131 & \bf 0.112 \\
\hline
    \end{tabular}}
    \caption{The few-shot learning retrieval performance of different methods on three  benchmark datasets. NDCG and ERR are at cut-off 20. }
    \label{tab:fewshot}
\end{table}

\begin{figure*}[!t]
    \centering
    \begin{minipage}[t]{.31\linewidth}
        \includegraphics[width=\linewidth]{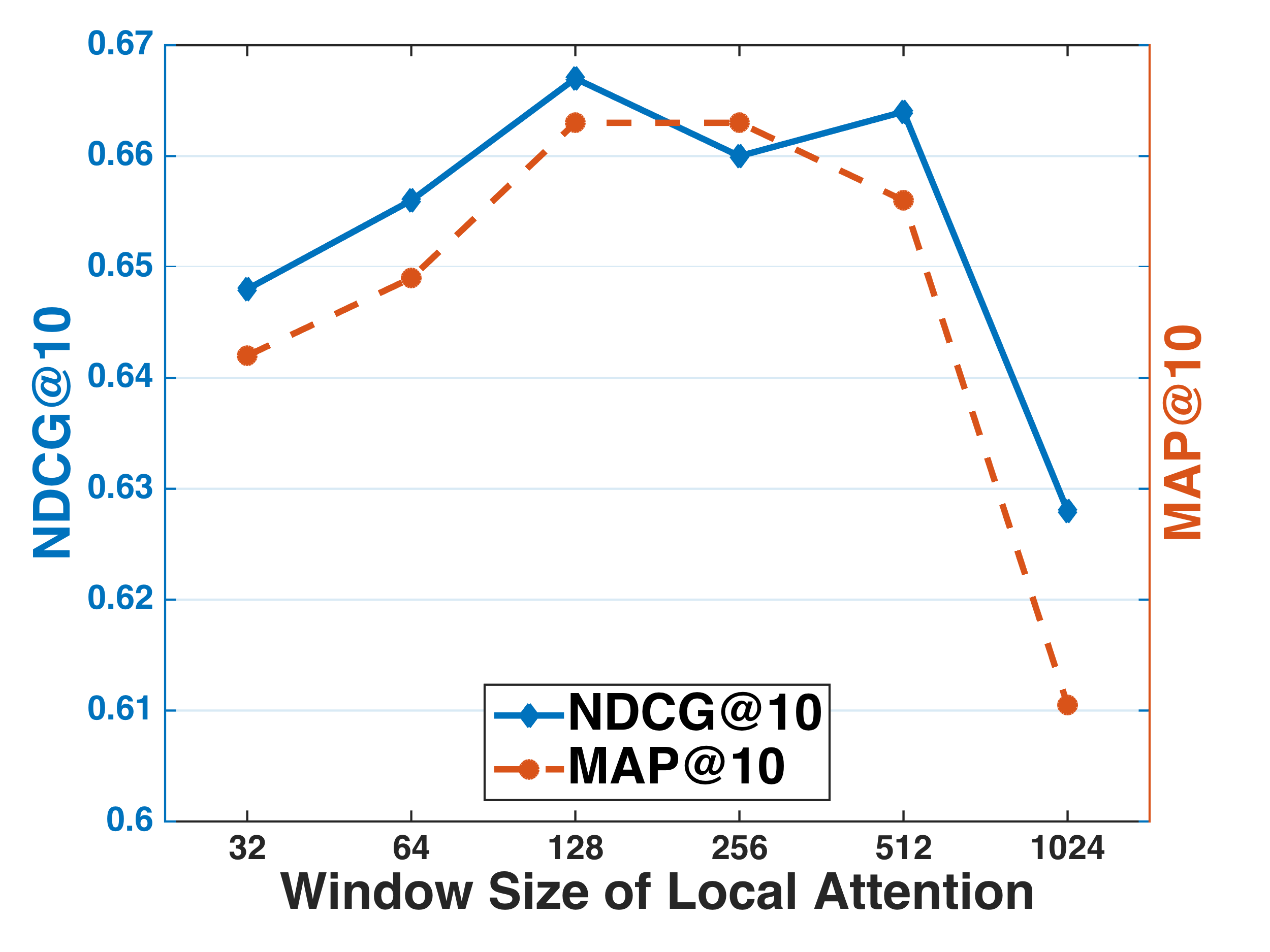}
        \caption{The performance of \modelname on TREC-19 DL track dataset with different local attention window sizes $w$.}
        \label{fig:windowsize}
    \end{minipage}
    \begin{minipage}{.01\linewidth}~\end{minipage}
    \begin{minipage}[t]{.31\linewidth}
        \includegraphics[width=\linewidth]{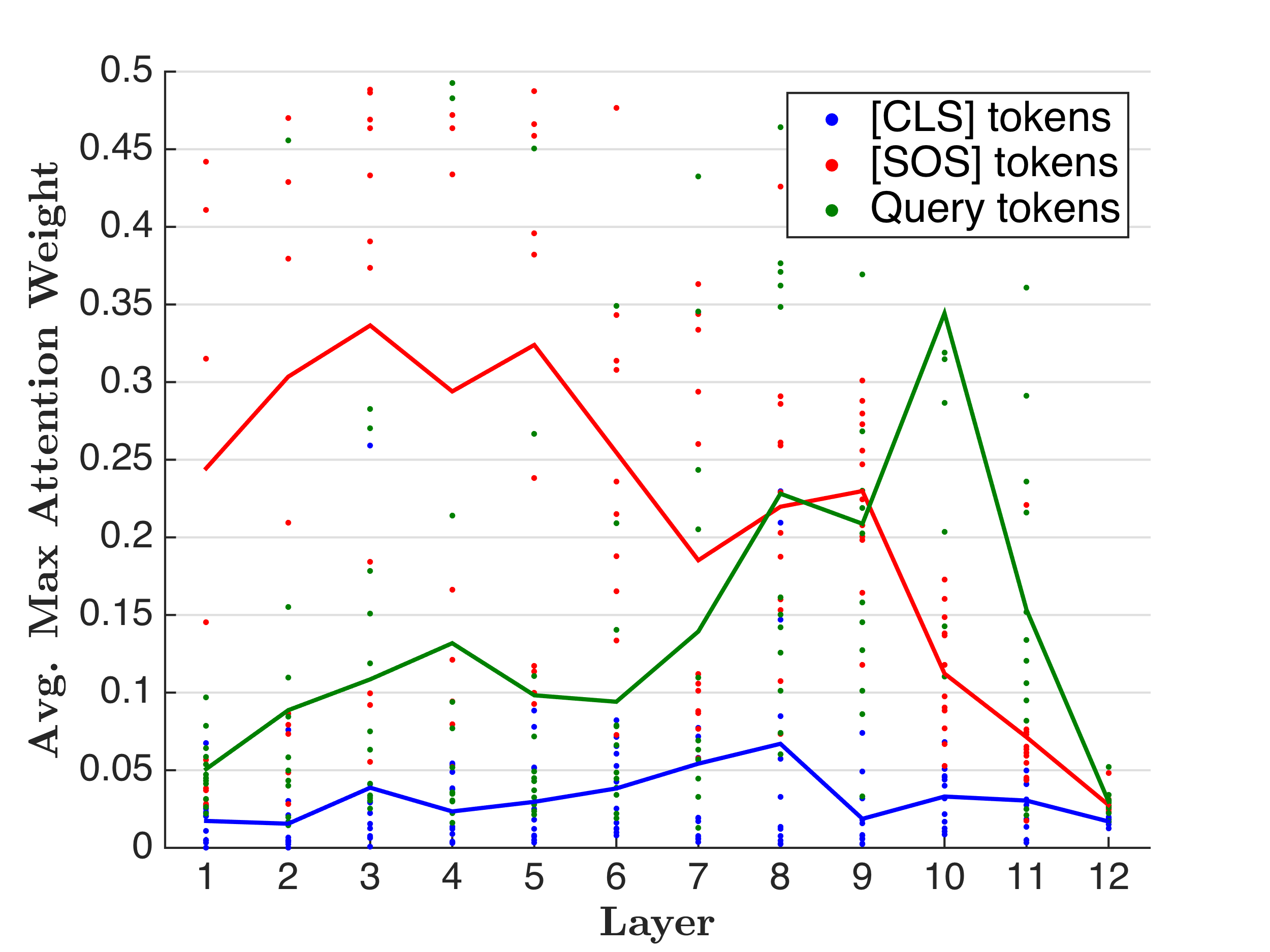}
        \caption{The average maximum attention scores to different types of tokens over Transformer layers. 
        }
        \label{fig:attn_weights}
    \end{minipage}
    \begin{minipage}{.01\linewidth}~\end{minipage}
    \begin{minipage}[t]{.31\linewidth}
        \centering
        \includegraphics[width=\linewidth]{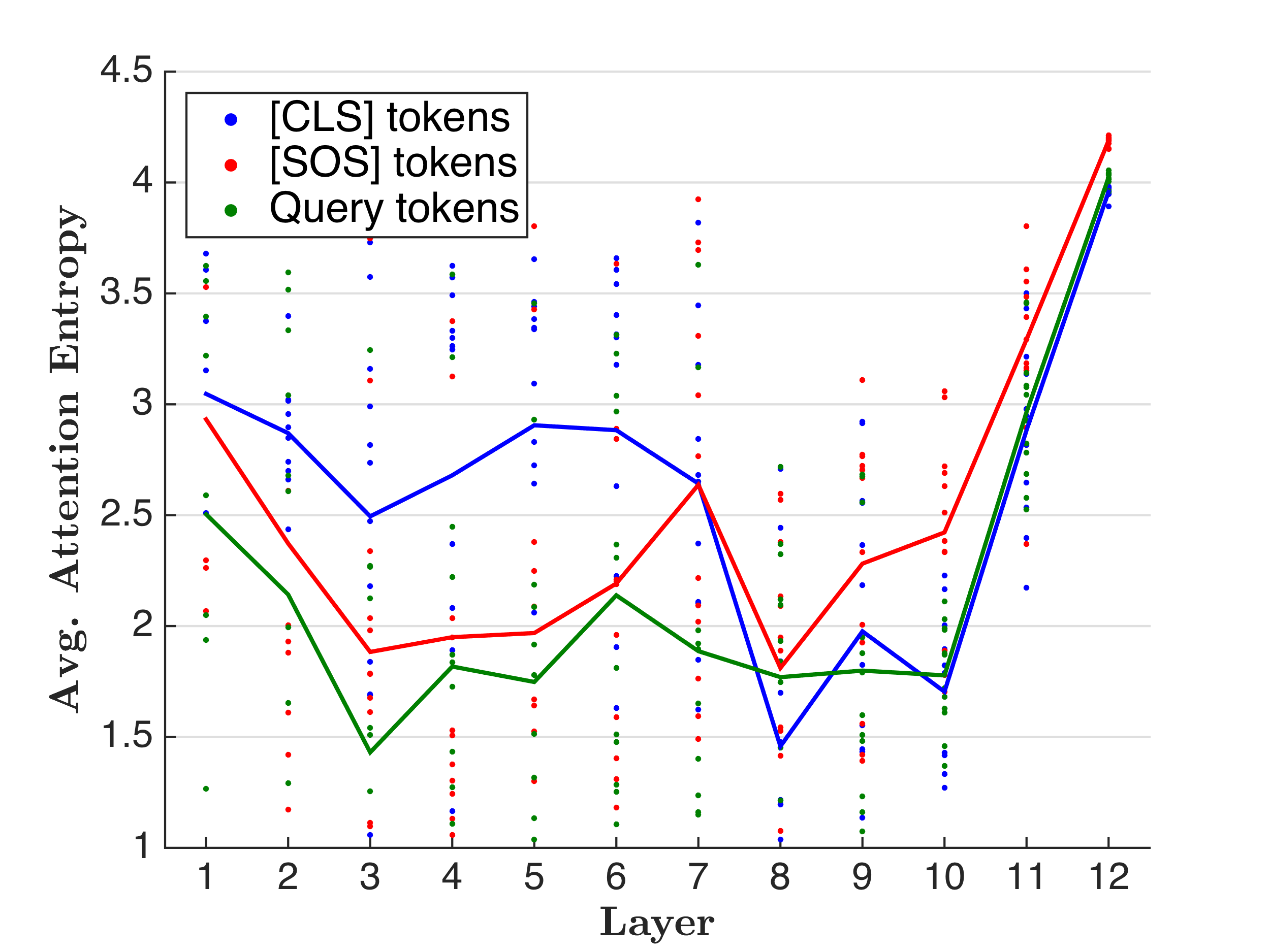}
        \caption{The average entropy scores of attention distributions for different token types over Transformer layers.}
        \label{fig:attn_entropy}
    \end{minipage}
\end{figure*}

\begin{table}[!t]
    \centering
    \resizebox{\linewidth}{!}{
    \begin{tabular}{l|c|c|ccc} \hline
     \multirow{2}{*}{Method} & \multicolumn{2}{c|}{Attention} & \multicolumn{2}{c}{TREC-19 DL Track} \\ \cline{2-5}
     & Q & Sent & NDCG@10 & MAP \\ \hline
RoBERTa (MaxP) & \cmark & \xmark & 0.630 & 0.246 \\
Sparse Transformer & \xmark & \xmark & 0.634 & 0.257 \\
LongFormer-QA & \xmark & \xmark & 0.627 & 0.255 \\
Transformer-XH & \cmark & \cmark & 0.646 & 0.256 \\
\hline
\modelname (S) & \xmark & \cmark & 0.633 & 0.244 \\
\modelname (Q) & \cmark & \xmark & 0.658 & 0.263 \\
\modelname & \cmark & \cmark & 0.667 & 0.278 \\

\hline
    \end{tabular}}
    \caption{The retrieval performance of different models on the TREC-19 DL track benchmark dataset with different global attention patterns.  Q and S indicate the usage of query and sentence global attention. Note that \modelname with no global attention is equivalent to Sparse-Transformer.}
    \label{tab:attnpattern}
    \vspace{-12pt}
\end{table}

\subsection{Effectiveness of Attention Patterns} 
This experiment studies the contribution of our query-directed sparse attention patterns to \qdst{}'s effectiveness.

Table~\ref{tab:attnpattern} shows the ablation results of the three attention patterns in TREC-19 DL benchmark: local attention only ($A_\text{local}$, Sparse Transformer), hierarchical attention on sentence only ($A_\text{sent}$, \qdst{} (S)), and query-oriented attention only ($A_\text{query}$, \qdst{} (Q)).
All three sparse attention patterns contribute. As expected, query-oriented attention is most effective to capture the relevance match between query and documents.
Note that the RoBERTa~(MaxP) and Transformer-XH also attend to queries, but the attention is more localized as the document is broke into separated text pieces and the query is concatenated with each of them.
In comparison, \qdst{} mimics the proximity matches and captures the global hierarchical structures in the document using dedicated attention from query terms to sentences.

Figure~\ref{fig:windowsize} depicts the change in retrieval effectiveness by varying the local attention window size.
Both NDCG@10 and MAP@10 grow at a steady pace starting from a window size of 32 and peak at 128, but no additional gain is observed with bigger window sizes.
The information from a term 512 tokens away does not provide many signals in relevance matching and is safely pruned in \qdst{}.
Note that the dip at attention size 1024 is because our model is initialized from RoBERTa which is only pretrained on 512 tokens.

\begin{table}[!t]
    \centering
    \resizebox{\linewidth}{!}{
    \begin{tabular}{l|rrrr} \hline 
    \multirow{2}{*}{Method} & \multirow{2}{*}{Length} & \multirow{2}{*}{Sparsity}
 & \multicolumn{2}{c}{ms per q-d} \\ \cline{4-5}
         & & &  Train & Infer\\ \hline
RoBERTa & 1024 & 100\% & 391 & 100 \\
RoBERTa & 2048 & 100\% & 799 & 205 \\
\hline
RoBERTa (FirstP) & 512 & 100\% & 138 & 17 \\
RoBERTa (MaxP)  & 4*512 & 25\% & 305 & 55 \\
Transformer-XH & 4*512 & 25\% & 309 & 54 \\
\hline
\modelname (128) & 512 & 30.84\% & 218 & 45 \\
\modelname (128) & 1024 & 18.72\% & 249 & 52 \\
\modelname (128) & 2048 & 8.97\% & 321 & 92 \\
\hline
Longformer-QA (128) & 2048 & 4.70\% & 166 & 45 \\
Sparse-Transformer  (128) & 2048 & 4.56\% & 154 & 40 \\
\modelname (32)  & 2048 & 6.70\% & 201 & 50 \\
\modelname (64) & 2048 & 8.97\% & 309 & 86 \\
\modelname (128) & 2048 & 13.53\% & 321 & 92 \\
\modelname (256) & 2048 & 22.64\% & 475 & 127 \\
\modelname (512)  & 2048 & 40.88\% & 512 & 160 \\
\modelname (1024)  & 2048 & 77.34\% & 629 & 195 \\
\hline
\modelname (Q) & 2048 & 5.10\% & 316 & 108 \\
\modelname (S) & 2048 & 8.57\% & 322 & 105 \\
\hline \multicolumn{5}{l}{\bf Without TVM Implementation} \\ 
Sparse-Transformer  (128)  & 2048 & 4.56\% & 251 & 62 \\
\modelname (128) & 2048 & 13.53\% & 390  & 103  \\
\hline
    \end{tabular}}
    \caption{Efficiency Quantification. The local attention window size is shown in parentheses.  Q and S indicate the usage of only query and sentence attention. Sparsity is compared with fully attention at same text length.} 
    \label{tab:efficiency}
\end{table}

\begin{table*}[!t]
    \centering 
    \scriptsize

     \resizebox{.95\linewidth}{!}{
    \begin{tabular}{p{.24\linewidth}p{.615\linewidth}} \hline 
        Q1: 1037798 (who is robert gray) & Q2: 1110199 (what is wifi vs bluetooth) \\  
        docid: D3533931  & docid: D1325409 \\ \hline 
        \hline
        \texttt{Heads 1,2,4,6,9,10,11,12:} & \texttt{Head {\color{white}0}1:} Bluetooth's low power consumption make it useful where power is limited.  \\ \cline{2-2}
        Robert Gray (title) & \texttt{Head {\color{white}0}2:} Wi-Fi appliances are often plugged into wall outlets to operate. \\ \hline
         \multirow{4}{\linewidth}{\texttt{Heads 3,5,7,8:} \newline Robert Gray, (born May 10, 1755, Tiverton, R.I. died summer 1806, at sea near eastern U.S. coast), captain of the first U.S. ship to circumnavigate the globe and explorer of the Columbia River.} & \texttt{Head {\color{white}0}7:} The extremely low power requirements of the latest Bluetooth 4.0 standard allows wireless connectivity to be added to devices powered only by watch batteries.  \\ \cline{2-2}
        & \texttt{Head {\color{white}0}9:} A Wi-Fi enabled network relies on a hub.  \\  \cline{2-2}
        & \texttt{Head 10:} The advantages of using bluetooth from existing technology. \\ \cline{2-2}
        & \texttt{Head 11:} Wi-Fi is more suited to data-intensive activities such as streaming high-definition movies, while Bluetooth is better suited to tasks such as transferring keyboard strokes to a computer.  \\ \cline{2-2}
        & \texttt{Head 12:} The greater power of Wi-Fi network also means it can move data more quickly than Bluetooth network. \\
        \hline
    \end{tabular}}

    \caption{Case study of two queries on the sentences with the highest attention weights in the last transformer layer over different heads for the [CLS] token.} 
    \label{tab:casestudy}
\end{table*}

\begin{table*}[!t]
    \centering
    \small 
    \resizebox{\linewidth}{!}{
    \begin{tabular}{lp{.9\linewidth}}\hline
    \multicolumn{2}{l}{Q3: 1112341 (what is the daily life of thai people)} \\ \hline
    Query Token &  Sentence with the highest attention weight in the document D1641978\\ \hline
    life & Children are expected to show great respect for their parents, and they maintain close ties, even well into adulthood . \\
    thai & Culture of Thailand (title) \\ \hline
    \end{tabular}}
    \caption{
    Case study of the query 1112341 on the sentences in the document D1641978 with the highest attention weights among all heads from two query tokens. Note that we use attention weights in the third transformer layer.}
    \label{tab:casestudy2}
\end{table*}

\subsection{Model Efficiency}
This experiment benchmarks the efficiency of different sparse attention patterns. Their training and inference time (ms per query-document pair, or MSpP) is shown in  Table~\ref{tab:efficiency}.

RoBERTa on 2048 tokens is prohibitive; we only measured its time with random parameters as we were not able to actually train it. Retrofitting was a natural choice to leverage pretrained models.

Sparsity helps. Sparse-Transformer (128) is much faster than MaxP. Interestingly, its attention matrix with only 4.56\% non-zero entries leads to on par efficiency with retrofitted solutions and also only 5 times faster compared to full attention; this is due to the required cost involved in feed-forward.
This effect is also reflected in the efficiency of \qdst{} with different local window sizes.

Different sparsity patterns dramatically influence the optimization of TVM. Intuitively, patterns with more regular shape would be easier to optimize than more customized connections in TVM. For example, the skipping patterns along sentence boundary in \qdst{} (S) seems more forgiving than the query-oriented attentions (Q). 
Comparing efficiency with and without our TVM implementation, the diagonal sparse shape in Sparse-Transformer is much better optimized.

How to better utilize the advantage from sparsity and structural inductive biases is perhaps a necessary future research direction
in an era where models with fewer than one billion parameters are no longer considered large~\cite{brown2020language}.
Making progress in this direction may need more close collaborations between experts in application, modeling, and infrastructure.

\subsection{Learned Attention Weights}
This experiment analyzes the learned attention weights in \qdst{}, using the approach developed by \citet{clark2019bertattention}. 

Figure~\ref{fig:attn_weights} illustrates the average maximum attention weights of the three attention patterns used in our model. Interestingly, the model tends to implicitly conduct hierarchical attention learning~\cite{yang2016hierarchical}, where lower layers focus on learning structures and pay more attention to [SOS] tokens, while higher layers emphasize the relevance by attending to queries more. Attention on both types of tokens is consistently stronger than on the [CLS] token. The model is capturing the inductive biases emphasized by our sparse attention structures. 

Figure~\ref{fig:attn_entropy} shows the average entropy of the attention weight distribution. 
Intuitively, lower layer attention tends to have high entropy and thus a very broad view over many words, to create contextualized representations. 
The entropy of query and [SOS] are in general lower, as they focus on capturing information needs and document structures.
The entropy of all three types of tokens rises again in the last layer, implying that they may try to aggregate representation for the whole input.

\subsection{Case~Study~on~Learned~Attention~Weights}
\label{section:case}

Table~\ref{tab:casestudy} shows a case study of sentences with the highest attention weight from [CLS] in the last layer for two example queries. For factoid query Q1, all heads center on precise sentences that can directly answer the query. For Q2 that is on the exploratory side, different attention heads exhibit diverse patterns focusing on partial evidence that can provide a broader understanding collectively.

Table~\ref{tab:casestudy2} depicts the other case study on learned attention weights of sentences from query tokens.
We adopt the third transformer layer, where sentences obtain more attention as shown in Figure~\ref{fig:attn_weights}, to emphasize significant sentences for query tokens.
The results show query-directed attention can capture sentences with different topics matched to individual query tokens, thereby comprehending sophisticated document structure.

These findings suggest that \qdst{} has an interesting potential to be applied to not only retrieval but also the question-answering task in NLP, providing a generic and effective framework, while also being interpretable with its the sparse structural attention connectivity.
We further provide an additional case study in Appendix~\ref{appendix:case}.

\section{Conclusions}
\label{section:conclusions}
\qdst{} improves the efficiency and effectiveness of pretrained transformers in long document ranking using sparse attention structures.
The sparsity is designed to capture the principal properties (IR-Axioms) that are crucial for relevance modeling: local contextualization, document structures, and query-focused matching. 
In four TREC document ranking tasks with variant settings,  \modelname consistently outperforms competitive baselines that retrofit to BERT or use sparse attention not designed for document ranking. 

Our experiments demonstrate the promising future of joint optimization of structural domain knowledge and efficiency from sparsity, while its current form is somewhat at the infancy stage.
Our analyses also indicate the potential of better interpretability from sparse structures and more unified models for IR and QA.

\bibliographystyle{acl_natbib}
\bibliography{emnlp2020}

\clearpage

\setcounter{page}{1}

\renewcommand{\appendixpagename}{\centering Appendix}
\begin{appendices}
\section{Experimental Details}
\label{appendix:expdetails}

In this section, we clarify the details about experimental datasets and experimental settings.

\subsection{Experimental Datasets}
\label{appendix:datasets}

\mysection{TREC-19 DL Track Dataset}
For ad-hoc retrieval, we adopt the TREC-19 DL track benchmark as the experimental dataset with training, dev, and test sets.
Training and dev sets consist of large-scale human relevance assessments derived from the MS MARCO collection~\cite{bajaj2016ms} with no negative labels and sparse positive labels for each query while relevance judgments in the test sets are annotated by NIST judges.

\mysection{Few-shot Document Ranking Benchmarks}
For few-shot learning, three retrieval benchmark datasets are utilized in our experiments, including Robust04, ClueWeb09-B, and ClueWeb12-B13.
Robust04 provides 249 queries from TREC Robust track 2014 with relevance labels.
ClueWeb09-B includes of 200 queries with relevance labels from TREC Web Track 2009-2012.
ClueWeb12-B13 consists of 100 queries from TREC Web Track 2013-2014 with relevance labels.

Note that Table~\ref{tab:datasets} in the paper summarizes the statistics of four experimental datasets.
Datasets of all benchmarks are publicly available.
The TREC-19 DL track provides all dataset on its offical website\footnote{https://microsoft.github.io/TREC-2019-Deep-Learning/}.
The queries and relevance assessments of three few-shot document ranking datasets can be found at the TREC website\footnote{https://trec.nist.gov/} while document collocations are also publicly available on the corresponding sites\footnote{RB04: https://trec.nist.gov/data/qa/T8\_QAdata/disks4\_5.html}\footnote{CW09: http://lemurproject.org/clueweb09/}\footnote{CW12: https://lemurproject.org/clueweb12/}.

\subsection{Experimental Settings}
\label{appendix:settings}

\mysection{Ad-hoc Retrieval}
Experiments follow the protocol of the TREC-19 deep learning track.
Each method is trained with the training set.
The model parameters can be further fine-tuned with the dev set and the MRR@10 metric.
The fine-tuned model is finally applied to the test set for evaluation.
Following the official metrics, MRR@10 is used in dev set runs as labels are incomplete and shallow, while the test set is comprehensively evaluated using NDCG@10 and MAP@10.

\mysection{Few-shot Document Ranking}
All experimental settings for few-shot learning are consistent with the``MS MARCO Human Labels'' setting in previous studies~\cite{zhang2020selective}.
Each method first trains a neural ranker on MARCO training labels, which are identical as in the TREC DL track.
The latent representations of trained models are then considered as features for a Coor-Ascent ranker for low-label datasets using five-fold cross-validation~\cite{dai2019deeper, conv_knrm} to rerank top-100 SDM retrieved results~\cite{metzler2007linear}.
Standard metrics NDCG@20 and ERR@20 are used to compare the different approaches. The results are reported by taking the average of each test fold from the total 5 folds, wherein the rest 4 folds in each round are used as training and dev queries.

\begin{table}[!t]
    \centering
    \resizebox{\linewidth}{!}{
    \begin{tabular}{cc|cc}\hline
        Method &  \#Params & Method &  \#Params\\\hline
        RoBERTa (FirstP) &  124M  & RoBERTa (MaxP) & 124M \\
        Sparse-Transformer & 149M  & Longformer-QA & 149M\\
        Transformer-XH & 128M & \modelname &  149M \\ \hline
    \end{tabular}}
    \caption{Number of parameters for methods.}
    \label{tab:num_params}
    \vspace{-12pt}
\end{table}

\begin{table*}[!t]
    \centering
    \small 
    \resizebox{.95\linewidth}{!}{
    \begin{tabular}{p{.9\linewidth}|r}\hline
    Sentence in the document D2944963 for Q4: 833860  (what is the most popular food in switzerland) & Top Query Token \\ \hline
    Top 10 Swiss foods with recipes (title) &  switzerland\\
    You certainly won't go hungry in Switzerland. & food \\ 
    You spear small cubes of bread onto long-stemmed forks and dip them into the hot cheese (taking care not to lose the bread in the fondue). & food \\ 
    Jamie Oliver has this easy cheese fondue recipe, and this five-star recipe has good reviews. & popular\\ \hline
    \end{tabular}}
    \caption{Case study of the query 833860 with the query tokens with the highest attention weights in the 10-th transformer layer among all heads from the [SOS] tokens of sentences in the document D2944963.}
    \label{tab:casestudy_from_sos}
\end{table*}

\mysection{Hyperparameter Settings and Search}
We adopt the pretrained model for sparse attention~\cite{Beltagy2020Longformer} and fix all of the hidden dimension numbers as 768 and the number of transformer layers as 12.
BERT-based models use RoBERTa as pretrained models~\cite{liu2019roberta}.
To hyperparameter tuning, we search the local attention window size $w$ in \{32, 64, 128, 256, 512, 1024\} with the dev set and determine $w=128$.
Models are optimized by the Adam optimizer~\cite{kingma2014adam} with a learning rate $10^{-5}$, $(\beta_1, \beta_2) = (0.9, 0.999)$, and a dropout rate 0.1.
Under the hyperparameter settings, the parameter numbers of our implemented methods are shown in Table~\ref{tab:num_params} summarizing the sizes of parameters based on \texttt{model.parameters()} in PyTorch.

\subsection{Evaluation Scripts}

All evaluation measures are computed by the official scripts.
For ad-hoc retreival, we use \text{trec\_eval}\footnote{https://github.com/usnistgov/trec\_eval} as the standard tool in the TREC community for evaluating ad-hoc retreival runs.
This is also the official setting of the TREC-19 deep learning track.
For few-shot document ranking, we use graded relevance assessment script (gdeval)\footnote{https://trec.nist.gov/data/web/10/gdeval.pl} as the evaluation script measuring NDCG and ERR.
Note that this setting is consistent with previous studies~\cite{zhang2020selective,dai2019deeper}.

\section{Baseline Methods}
\label{appendix:baselines}

In this section, we introduce each baseline method.

\mysection{TREC Best Runs}
\begin{itemize}
    \item \textbf{bm25tuned\_prf}~\cite{yang2019reproducing} fine-tunes the BM25 parameters with pseudo relevance feedback as the best BM25 based method in official runs.
    \item \textbf{srchvrs\_run1} is marked as the best traditional ranking method among official runs ~\cite{craswell2020overview}.
    \item \textbf{TUW19-d3-re}~\cite{hofstatter2019tu} as the best method without using non-pretrained language models (non-PLM) in official runs utilizes a transformer to encode both of the query and the document, thereby measuring interactions between terms and scoring the relevance.
    \item \textbf{bm25\_expmarcomb}~\cite{yilmaz2019h2oloo} combines sentence-level and document-level relevance scores with a pretrained BERT model.
\end{itemize}

\mysection{Classical IR Methods}
\begin{itemize}
    \item \textbf{SDM}~\cite{metzler2005markov} as a sequential dependence model conducts ranking based on the theory of probabilistic graphical models. We obtain ranking results of SDM from previous studies~\cite{dai2019deeper}. SDM is not only treated as a baseline method but also providing the candidate documents for reranking in the few-shot learning task.
    \item \textbf{Coor-Ascent}~\cite{metzler2007linear} is a linear feature-based model for ranking. It is also considered as the trainer in few-shot learning with representations from methods.
\end{itemize}

\mysection{Neural IR Methods}
\begin{itemize}
    \item \textbf{CO-PACRR}~\cite{hui2018co} utilizes CNNs to model query-document similarity matrices and provide a score using a max-pooling layer.
    \item \textbf{Conv-KNRM}~\cite{conv_knrm} applies CNNs to independently encode the query and the document. The encoded representations are then integrated by a cross-matching layer, thereby deriving relevance scores.
\end{itemize}

\mysection{Transformer-based Methods}
\begin{itemize}
    \item \textbf{TK}~\cite{hofstatter2020interpretable} and \textbf{TKL}~\cite{hofsttter2020local} apply transformers to independently model the query and document, thereby measuring term interactions at the embedding level.
    \item \textbf{RoBERTa (FirstP)} and \textbf{RoBERTa (MaxP)}~\cite{dai2019deeper} adapt long-form documents by considering the first paragraph and combining RoBERTa outputs with max-pooling over paragraphs. Note that each paragraph is also attached with query tokens before being fed into the model.
    \item \textbf{Transformer-XH}~\cite{zhao2020transformerxh} encodes each sentence independently and considers their relations with an extra-hop attention layer. Note each sentence is also attached with query tokens as the model input.
    \item \textbf{Sparse-Transformer}~\cite{child2019sparsetransformer} simply uses sparse local attention to tackle the efficiency issue of transformers.
    \item \textbf{Longformer-QA}~\cite{Beltagy2020Longformer} extends Sparse-Transformer by attaching two global attention tokens to the query and the document as their settings for question answering. Note that their global attention would not consider document structural information.
\end{itemize}

\section{Additional~Study~on~Attention~Weights}
\label{appendix:case}

In addition to attention from the classification token [CLS] and query tokens as shown in Section~\ref{section:case}, here we analyze the attention from sentences.
Table~\ref{tab:casestudy_from_sos} shows the query tokens with the highest attention weights in the 10-th transformer layer among all head from the [SOS] tokens of sentences.
Note that the 10-th transformer layer indicates higher importance of query tokens as shown in Figre~\ref{fig:attn_weights}.
The results show that \modelname is capable of directing sentences to the tokens with matched topics, thereby understanding sophisticated document structure with different topics.

\end{appendices}

\end{document}